\documentclass[letterpaper, 10 pt, conference]{ieeeconf}  

\IEEEoverridecommandlockouts                              

\usepackage{amsfonts} 
\usepackage{algorithm}
\usepackage{cite}
\usepackage{graphics} 
\usepackage{graphicx}
\usepackage{subcaption}
\usepackage{tabularx}
\usepackage{epsfig} 
\usepackage{amsmath} 
\usepackage{textcmds}
\usepackage[dvipsnames]{xcolor}
\title{\LARGE \bf
SEIP: \textbf{S}imulation-based Design and \textbf{E}valuation of \\ \textbf{I}nfrastructure-based Collective \textbf{P}erception
}

\author{Ao Qu$^{1\dag}$, Xuhuan Huang$^{2\dag}$, Dajiang Suo$^{3*}$
\thanks{\dag Equal contribution}
\thanks{\ddag This work was partially supported by MIT Mobility Initiative}
\thanks{$^{1}$Ao Qu is with the Department of Civil and Environmental Engineering at Massachusetts Institute of Technology, Cambridge, MA 02139 USA
        {\tt\small qua@mit.edu}}%
\thanks{$^{2}$Xuhuan Huang is with Spatial Sciences Institute at University of Southern California,
        Los Angeles, CA 90089
        {\tt\small xuhuanhu@usc.edu}%
}
\thanks{$^{3}$Dajiang Suo is with the Department of Mechanical Engineering at Massachusetts Institute of Technology, Cambridge, MA 02139 USA
        {\tt\small djsuo@mit.edu}%
}
\thanks{ Corresponding author: Dajiang Suo}}

\begin{document}

\maketitle
\thispagestyle{empty}
\pagestyle{empty}

\begin{abstract}

Recent advances in sensing and communication have paved the way for collective perception in traffic management, with real-time data sharing among multiple entities. While vehicle-based collective perception has gained traction, infrastructure-based approaches, which entail the real-time sharing and merging of sensing data from different roadside sensors for object detection, grapple with challenges in placement strategy and high ex-post evaluation costs. Despite anecdotal evidence of their effectiveness, many current deployments rely on engineering heuristics and face budget constraints that limit post-deployment adjustments. This paper introduces polynomial-time heuristic algorithms and a simulation tool for the ex-ante evaluation of infrastructure sensor deployment. By modeling it as an integer programming problem, we guide decisions on sensor locations, heights, and configurations to harmonize cost, installation constraints, and coverage. Our simulation engine, integrated with open-source urban driving simulators, enables us to evaluate the effectiveness of each sensor deployment solution through the lens of object detection. A case study with infrastructure LiDARs revealed that the incremental benefit derived from integrating additional low-resolution LiDARs could surpass that of incorporating more high-resolution ones. The results reinforce the necessity of investigating the cost-performance tradeoff prior to deployment. The code for our simulation experiments can be found at \textcolor{blue}{https://github.com/dajiangsuo/SEIP}.

\end{abstract}

\section{INTRODUCTION}
Advances in sensing and communication technologies have made possible the real-time sharing and merging of data collected by multiple traffic entities for collective perception. When a vehicle or infrastructure sensor detects an anomaly or obstacle in its vicinity, it disseminates this information to other entities in real time. Consequently, even vehicles or infrastructure that may not have direct line-of-sight or sensory access to the obstacle become cognizant of its presence. Therefore, this approach helps individual traffic participants form a bird's view of the target region and prevents occlusion risks in urban areas with high traffic volume. While the vehicle-based collective perception has been well explored, the infrastructure-based collective perception, which requires the deployment of multiple roadside sensors, has been hindered by the lack of guidance for roadside sensor placement and the potentially high costs for ex-post evaluation. For sensors with high procurement and installation costs, such as light detection and ranging (LiDAR), there is an urgent need for effective ways of ex-ante design and evaluation.


Decisions on infrastructure-sensor deployment strategies are often made based on engineering heuristics. These decisions require balancing factors such as sensor resolution, range, number, cost, locations, and placement height. Optimizing the configurations and placements of infrastructure sensors is crucial to minimize occlusion risks. However, for projects constrained by budget, it might not be feasible to make ex-post adjustments to deployment strategies.

This paper presents algorithms and a simulation-based tool to support the ex-ante assessment of infrastructure sensor deployment for collective perception applications. The infrastructure-sensor deployment is formulated as an integer programming problem. It can be solved with polynomial-time heuristics to derive sensor locations, installation heights, and configurations. The goal is to balance procurement cost, physical constraints for installation, and sensing coverage (i.e., to what extent a target region is monitored by the sensors installed). Additionally, we implement the proposed integer programming algorithms in a sensor-deployment engine to support the evaluation process. In addition to interacting with open-source urban driving simulators for essential sensor deployment data, the engine can extract and merge synthetic data from various infrastructure sensors synchronously. This allows us to evaluate the effectiveness of each sensor deployment solution through the lens of object detection, one of the critical tasks in traffic monitoring. The application of the proposed framework was illustrated through a case study on traffic monitoring by using infrastructure LiDARs. Results suggest that, for projects with a limited sensing budget, the solution combining multiple low-resolution LiDARs (i.e., low resolution or small range) is not necessarily inferior to the solution with fewer high-resolution LiDARs. The results also highlight the need to investigate the cost-performance tradeoff before the actual deployment.

\section{Previous work}
Vehicle-based collective sensing solutions have been well-explored in~\cite{chen2019f,chen2019cooper,marvasti2020cooperative,aoki2020cooperative,xu2022opv2v,wang2020v2vnet}. Also, there exists work on monitoring traffic conditions in a given region with the support of infrastructure sensors. 

Chtourou et al. found through vehicular simulation that infrastructure-assisted collective perception can result in an eight-time higher number of recognized objects than only relying on vehicle-assisted collective perception~\cite{chtourou2021collective}. In~\cite{li2022v2x}, the authors evaluated the performance improvement in identifying traffic participants near the intersection with the assistance by infrastructure sensors that locate at a fixed position. Specifically, the infrastructure LiDAR was arbitrarily placed at the center of an intersection to provide V2I collective sensing services to nearby vehicles. Building on simulation, the author in~\cite{mao2022dolphins} constructed datasets containing sensing data from both vehicles and infrastructures in six safety-critical traffic scenarios.

There also exists work on deploying infrastructure sensors to roadside units in the real world for collective perception applications. For example, Ye et al. placed roadside cameras near intersections and collected sensing data from both infrastructure and vehicle-mounted sensors to construct a real-world collective perception dataset~\cite{ye2022rope3d}. The roadside cameras were attached to the pole near intersections to capture image data. This infrastructure-sensing data was then fused with vehicle-mounted LiDAR for object detection tasks. In~\cite{yu2022dair}, high-end LiDARs were mounted on each of the four traffic lights near each intersection to measure the surrounding traffic conditions. 

To provide guidance for engineers to determine the infrastructure sensor deployment strategies that achieve the tradeoff between cost and performance, we extend prior research in algorithmic and simulation-based approaches to infrastructure-sensor deployment~\cite{borger2022sensor}, primarily related LiDAR sensors~\cite{cai2022analyzing,borger2022sensor,kloeker2022generic}. To the best of our knowledge, the most closely related work to ours is given in~\cite{cai2022analyzing,vijay2021optimal}.

The author in~\cite{cai2022analyzing} has contributed to infrastructure sensor deployment by developing a simulation library to represent the physical characteristics of different LiDAR sensors. Although the simulation-based pipeline can significantly reduce the evaluation cost of infrastructure LiDAR deployment, the selection of the exact sensor deployment locations in~\cite{cai2022analyzing} is still arbitrary without any guidance from algorithmic approaches. In particular, the cost-performance tradeoff is not explicitly explored. The work in~\cite{vijay2021optimal} proposes a novel algorithm based on integer programming formulation for infrastructure sensor deployment. However, it only considers sensing coverage or LiDAR occupation (i.e., the area in target objects that are visible to sensors), neglecting the actual traffic flow and how well LiDAR can help object detection, the key task in which infrastructure sensors are used. Additionally, none of them considers different levels of importance among regions of interest under sensing budget constraints.  

\section{An integer programming formulation of infra-sensor deployment}

\subsection{Sensor visibility and road discretization}
Since the ultimate goal of adopting infrastructure-based cooperative perception is to reduce occlusion risks in perceiving objects in regions of interest, it is necessary to derive information as to whether a given region can be \qq{seen} by sensors deployed on the roadside. This is often referred to as sensor visibility and coverage~\cite{kloeker2022generic,vijay2021optimal} or \qq{occupancy} for LiDARs~\cite{kim2019placement,berens2021genetic}. The ideal configuration (e.g., low resolution vs. high resolution) and placement of sensors is expected to monitor as much of that region as possible with high accuracy.

\begin{figure}[tb!]
\centering
\includegraphics[width=1\columnwidth]{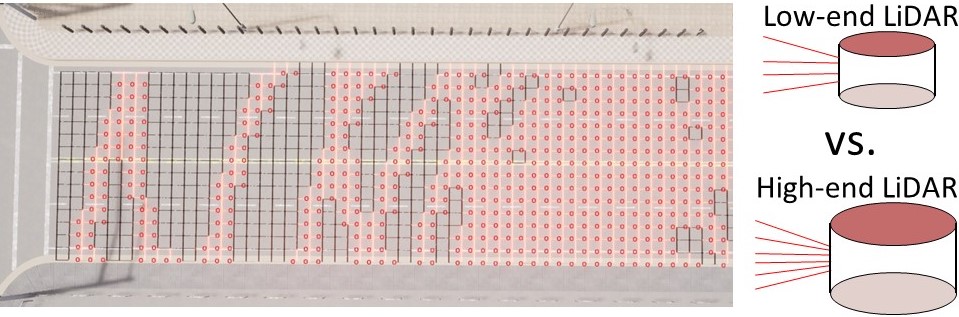}
\caption{An example of road discretization and visibility grids.}
\label{fig:vis_grid_demo}
\end{figure}

To reduce the complexity brought by determining sensor visibility in continuous regions of interest (ROI), we first discretize each road segment into grids, as shown in Fig.~\ref{fig:vis_grid_demo}. To put it formally, if geometries of road segments are captured by high-definition maps, we can then represent an ROI $R$ as a 2-dimensional plane $R = \cup R_i$ where each $R_i$ is a road segment. To determine sensor locations and quantify relevant measurements, we define sensor candidate points $S = \{s_1, \dots, s_{N_S}\}$ and target points $T = \{t_1, \dots, t_{N_T}\} \subset R$.

Additionally, we find the minimum and maximum values in each of the two dimensions of $R$, denoted as $x_{min}, x_{max}, y_{min}, y_{max}$. Then, we find uniformly spaced coordinates $P = \{(x_i, y_j) | x_{0} > x_{min}, x_N < x_{max}, y_{0} > y_{min}, y_M < y_{max}, i\in 1, \dots, N, j\in 1, \dots, M\}$ and $T$ can be determined as $\{(x_i, y_j)| \exists R_k, (x_i, y_j) \in R_k\}$. In practice, this process can be accelerated by spatial searching algorithms such as $R$-tree.

After each road segment in the ROI becomes discretized, we can derive sensor visibility in a more tractable way. To follow the conventions and adapt notations in previous work on sensor visibility, we construct visibility grid $V \in \mathbb{R}^{N_S \times N_T}$ in which $V_{i, j} = 1$ if the sensor $s_i$ is able to \qq{see} the target point $t_j$ and $V_{i, j} = 0$ otherwise, as shown in Fig.~\ref{fig:vis_grid_demo}. The grids with a red dot indicate that they are covered by at least one sensor. To determine the value of $V_{i, j}$, we leverage the formula given in~\ref{equ::visibility}.

\begin{equation}
V_{i, j} = \begin{cases}
    &1 \text{ if } \exists d\in D_i, \texttt{dist}(d, t_j) < \delta \\ 
    &0 \text{ otherwise} \label{equ::visibility}\\
\end{cases}
\end{equation}

$d=(x,y,z,i)$ in eq.~\ref{equ::visibility} represents a point received by LiDARS $(x,y,z)$ are the 3-dimensional coordinates and $i$ is the intensity. $D_i$ is the ray-casting data for sensor $s_i$. We require that the distance between $d$ and the target point $t_j$ be less than a threshold to ensure full coverage for the ROI (i.e., $V_{i, j} = 1$). Similarly, additional requirements on the intensity level $i$ can be added to eq.~\ref{equ::visibility}. 

The equation for determining sensor visibility was denoted implicitly in previous work without explaining the underlying rationales. Our interpretation is that the criterion presented in eq.~\ref{equ::visibility} can be appreciated by the voxelization operation, a classic technique used to encode point-cloud data by storing them into regular grids, also known as voxels. 

Consider \textit{PointPillars} \cite{Lang_2019}, the state-of-the-art deep neural network for 3D object detection based on point cloud. To save memory and accelerate the processing of 3D dimensional data, an encoder network is included in \textit{PointPillars} for point cloud data compression. Rather than feeding the neural networks with raw point cloud data, the encoder network will first discretize the \qq{cloud} into evenly-spaced grids in the x-y dimension to create a set of pillars (i.e., voxels with unlimited z dimension), which is consistent with the road discretization presented earlier. Pillars without any points will be discarded without further processing. 

Therefore, if the rectangular area that is represented by target point $t_j$ is \qq{hit} by the laser beams from sensor $s_i$, there is a high chance that the pillar corresponding to $t_j$ will be considered as valid and included by \textit{PointPillars} \cite{Lang_2019} for further processing.

For sensor candidate points, in our study, we assume that sensors can be placed anywhere on sidewalks and traffic signals. Having access to the geometry of each nearby sidewalk segment and locations of traffic signals, we can easily construct $S$ with a similar uniform discretization method. While this assumption may become unrealistic in some real-world scenarios, the same framework can be easily adapted.

\subsection{Sensor deployment algorithms}
After sensor visibility is constructed, we are ready to present our algorithms to derive sensor deployment strategies that determine the locations, heights, and configurations (e.g., resolution) of LiDAR sensors. Note that our objective here is to maximize sensing coverage (i.e., the target points seen by sensors) under budget and physical constraints that introduce additional requirements on sensor locations and installation heights, which differs from previous work~\cite{cai2022analyzing,vijay2021optimal}.

We address the importance of physical constraints as it helps stakeholders save deployment costs and conform with local regulations. For example, in the U.S., to utilize the power and communication components of existing digital infrastructure, it is recommended by the Department of Transportation that new roadside sensing and communication modules be collocated with signal poles or mast arms~\cite{learned2020connected}. Additionally, since most transportation construction projects are jointly funded by funds from federal, state, and local agencies, they must not only abide by federal laws but also conform with regulations defined by local transportation departments and even municipals. For example, many cities and counties in the U.S. have regulations regarding how and where renovation projects of roads may take place~\cite{cambridge_streetrule}.

The integer programming formulation for infrastructure sensor deployment is given in Algorithm~\ref{alg:bip} and~\ref{alg:bip_weight}, respectively. In Algorithm 1, each target point $t_j$ is treated equally. This is consistent with the ROI uniformity metric~\cite{cai2022analyzing}, which measures the uniformity of point cloud distribution across all target regions. On the other hand, the objective function in Algorithm 2 can ensure that higher weights be assigned to those areas that are more important within the ROI. 

\begin{figure*}[tb!]
\centering
\includegraphics[width=0.8\textwidth]{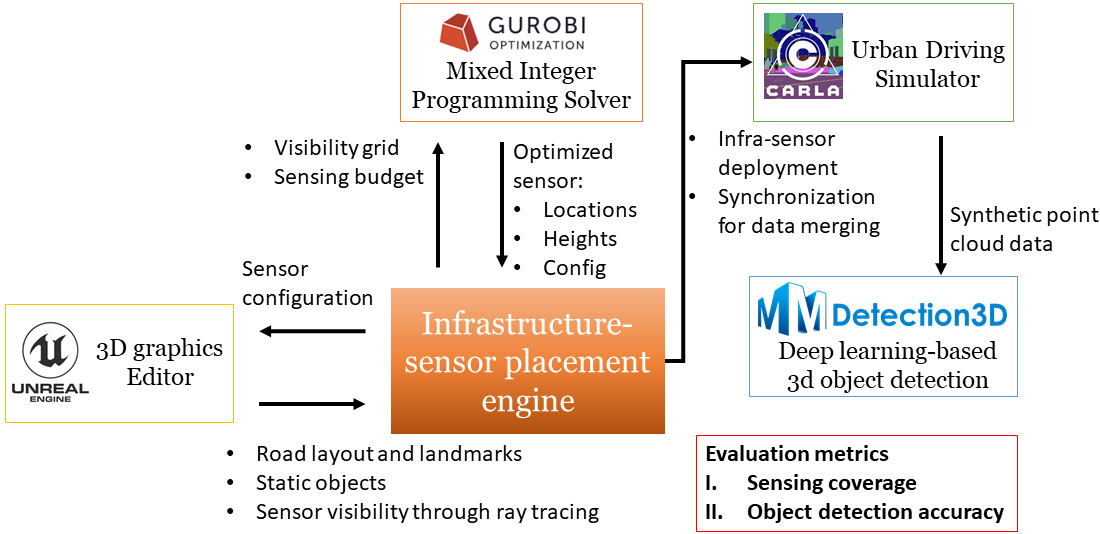}
\caption{The python-based infrastructure sensor deployment engine}
\label{fig:framework}
\end{figure*}

\begin{algorithm}
\caption{\textbf{Vanilla} Infra-sensor Deployment}\label{alg:bip}
\textbf{Require} Visibility grid $V\in \mathbb{R}^{N_S\times N_T}$, Sensing budget $N$ \\
\begin{align*}
\text{max} &\sum_{i}^{N_T} y_i \\
\text{subject to } &\sum_{i}^{N_S} c(x_i) \leq C\\
&\sum_{i}^{N_S} V_{i, j} \cdot x_i \geq y_i, \ \forall j \in 1, \dots, N_T \\
&x_i \in \{0, 1\},\ \forall i \in 1, \dots, N_S \\
&y_i \in \{0, 1\},\ \forall i \in 1, \dots, N_T
\end{align*}
\textbf{Return} $\{s_1, \dots, s_{N_S}\}$
\end{algorithm}

\begin{algorithm}
\caption{\textbf{Weighted} Infra-sensor Deployment}\label{alg:bip_weight}
\textbf{Require} Visibility grid $V\in \mathbb{R}^{N_S\times N_T}$, Sensing budget $N$ \\
\begin{align*}
\text{max}   &\sum_{i}^{N_T} y_i \cdot w_i \\
\text{subject to } &\sum_{i}^{N_S} c(x_i) \leq C\\
&\sum_{i}^{N_S} V_{i, j} \cdot x_i \geq y_i, \ \forall j \in 1, \dots, N_T \\
&x_i \in \{0, 1\},\ \forall i \in 1, \dots, N_S \\
&y_i \in \{0, 1\},\ \forall i \in 1, \dots, N_T
\end{align*}
\textbf{Return} $\{s_1, \dots, s_{N_S}\}$
\end{algorithm}

In both Algorithm~\ref{alg:bip} and~\ref{alg:bip_weight}, we use $\{x_i\}_{i=1}^{S_N}$ and $\{y_j\}_{j=1}^{S_T}$ as binary variables for sensor candidate points and target points. $x_i = 1$ denotes that sensor $s_i$ is selected, and vice versa. Similarly, $y_j = 1$ means the target point $t_j$ is visible to the selected sensor. Since we seek to maximize the sensing coverage, we want to maximize the number of target points that are visible to sensors.

The first constraint $\sum_{i}^{N_S} c(x_i) \leq C$ ensures that the sensor procurement and installation costs do not exceed the total budget $C$. For clarity in our upcoming case study in Section V, we introduce an alternative constraint $\sum_{i}^{N_S} x_i \leq N$, meaning the maximum number of sensors that can be used. This assumption operates under the premise that specific sensor configurations, such as resolution, are predetermined and the costs of individual sensor modules, like LiDAR, remain constant. However, in practical applications, the varied failure modes associated with each sensor type can lead to distinct maintenance costs, as detailed in~\cite{yu2020optimizing}, introducing further constraints to the optimization process 

The second constraint $\sum_{i}^{N_S} V_{i, j} \cdot x_i \geq y_i$ is used to guarantee that a region or a target point $t_j$ will not be regarded as \qq{visible} unless at least one selected sensor can actually see it.

\begin{figure*}[tb!]
\centering
\begin{subfigure}{0.45\textwidth}
  \centering
  \includegraphics[width=\textwidth]{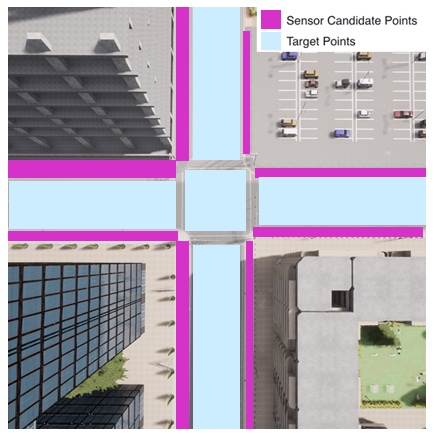}
  \caption{The bird's view of ROI and candidate installation locations.}
  \label{fig:birdview}
\end{subfigure}
\begin{subfigure}{0.45\textwidth}
  \centering
  \includegraphics[width=\textwidth]{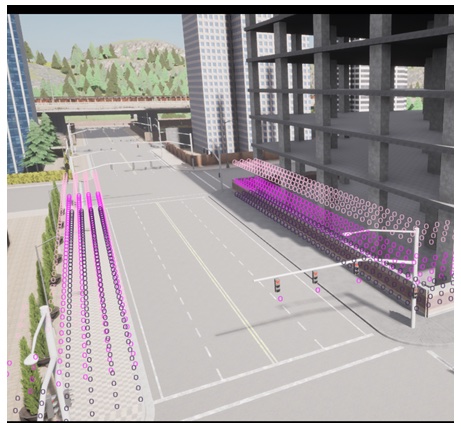}
  \caption{The candidate installation heights.}
  \label{fig:horizontalview}
\end{subfigure}
\caption{The digital traffic scenario used in our experiments.}
\label{fig:worldmap}
\end{figure*}

\section{A simulation-based infra-sensor \\ deployment engine}
To automate the evaluation pipeline of infrastructure-based collective perception, we implement an infrastructure-sensor deployment engine by using Python programming language. In addition to optimizing infrastructure sensor placement, the engine can also deploy infrastructure sensors within the urban driving simulator that we adopted for point cloud data generation, as shown in Fig.~\ref{fig:framework}.

As a prerequisite for optimizing Infrastructure LiDAR deployment, we need to construct the visibility grids discussed earlier (the inputs to Algorithms 1 and 2). This is because all the information about road geometries and layouts, landmarks such as signals, and static roadside objects might cause occlusions on the path of sensor signal propagation (e.g., LiDAR beams) and thus influence sensor visibility. 

Fortunately, there exist industry standards that enable the digitalization of physical objects in the real world into assets in digital maps, making the construction of the visibility grid possible. In our implementation, we leverage the digital maps defined based on the OpenDRIVE format, a standard defined by the Association for Standardisation of Automation and Measuring Systems (ASAM), to ease the editing and exchange of road networks for driving simulation. The detailed process of capturing geospatial and vision data from real-world traffic scenarios for building OpenDrive-compatible digital twin maps is out of the scope of this paper. Those interested in further exploration of this topic may refer to~\cite{zheng2022citysim}.

To automate the query of the information stored in the digital maps, we use \textit{Unreal Engine}~\cite{unreal}, a computer graphics editor for creating digital assets used in 3D virtual games. The APIs provided by the Unreal Engine also allow the infrastructure-sensor placement engine to conduct ray casting for each infrastructure LiDAR that we have deployed to the target traffic scenario. 

It should be noted that the visibility grid for each LiDAR sensor only considers the occlusions caused by static objects (e.g., buildings and trees) without accounting for dynamic objects such as moving vehicles. This implies the necessity of adopting a second evaluation criterion, \textit{object detection accuracy}, for infrastructure sensor deployment.

With all the inputs needed for Algorithms 1 and 2 ready, we run our proposed sensor deployment algorithms by using the Mixed Integer Programming Solver provided by the Gurobi Optimizer, a solver toolbox for mathematical programming. Although the problem of infrastructure sensor deployment is NP, greedy algorithms can achieve $1-\frac{1}{e}$ approximation within polynomial time, and many other heuristic algorithms have been proposed \cite{lan2007effective,vasko1984efficient}.

In addition to optimizing sensor deployment, the second role of the sensor deployment engine in the evaluation pipeline is to generate and synchronize synthetic sensing data from multiple infrastructure LiDARs for object detection. Here, we use CARLA~\cite{dosovitskiy2017carla}, an open-source urban driving simulator in 3D traffic scenarios. CARLA allows us to simulate vehicle movement in the target traffic scenario and precisely control the sensor-deployment locations and heights.

After we decide on sensor locations and create them in the urban driving simulator to generate synthetic data, we can leverage deep learning-based approaches for object detection, the second metric we use for evaluating collective perception performance, as shown in Fig.~\ref{fig:framework}. For this task, we utilize the MMDetection3D framework~\cite{mmdet3d2020}, which is an open-source platform that encapsulates state-of-the-art machine-learning models for 3D object detection.

\section{Experimental evaluation and discussion}


\begin{table}
\caption{The specifications of LiDARs in experiments}
\label{tab:lidar}
\setlength{\tabcolsep}{3pt}
\begin{tabular}{|p{51pt}|p{140pt}|p{40pt}|}
\hline 
Sensor type& 
Configuration settings & Cost range\\
\hline
Infrastructure LiDAR: Type 1& 16 channels, 5Hz capture frequency, 360$^{\circ}$ horizontal FOV, -15$^{\circ}$ to 15$^{\circ}$ vertical FOV, $\leq100m$ range, $\pm3cm$ accuracy, 300k points per second& $\$$4--8k\\
\hline
Infrastructure LiDAR: Type 2& 
32 channels, 5Hz capture frequency, 360$^{\circ}$ horizontal FOV, -25$^{\circ}$ to 15$^{\circ}$ vertical FOV, $\leq200m$ range, $\pm3cm$ accuracy, 1.2M points per second & $\$$10--20k \\
\hline
Infrastructure LiDAR: Type 3& 128 channels, 5Hz capture frequency, 360$^{\circ}$ horizontal FOV, -25$^{\circ}$ to 15$^{\circ}$ vertical FOV, $\leq300m$ range, $\pm3cm$ accuracy, 4.8M points per second& $\$$70--90k \\
\hline
\end{tabular}
\label{tab:sensor_config}
\end{table}

\begin{figure*}[tb!]
\centering
\includegraphics[width=0.9\textwidth]{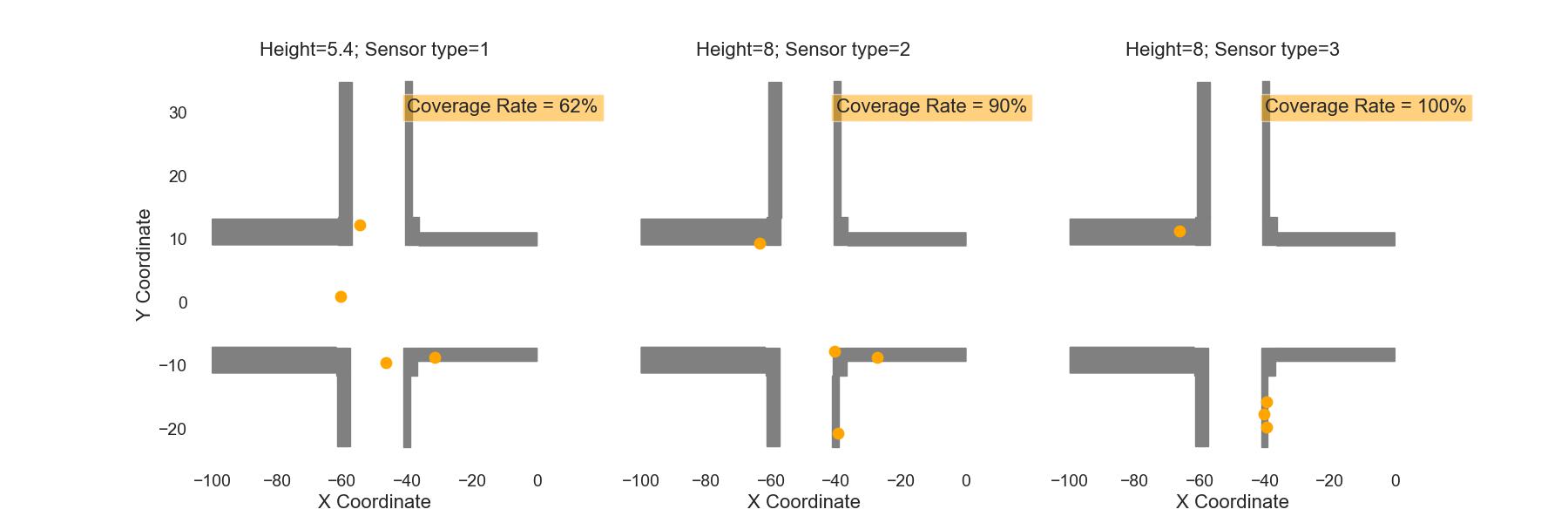}
\caption{The recommended sensor deployment locations and heights with four LiDARs}
\label{fig:recommended_loc}
\end{figure*}

\begin{figure*}[tb!]
\centering
\begin{subfigure}{0.32\textwidth}
  \centering
  \includegraphics[width=\textwidth]{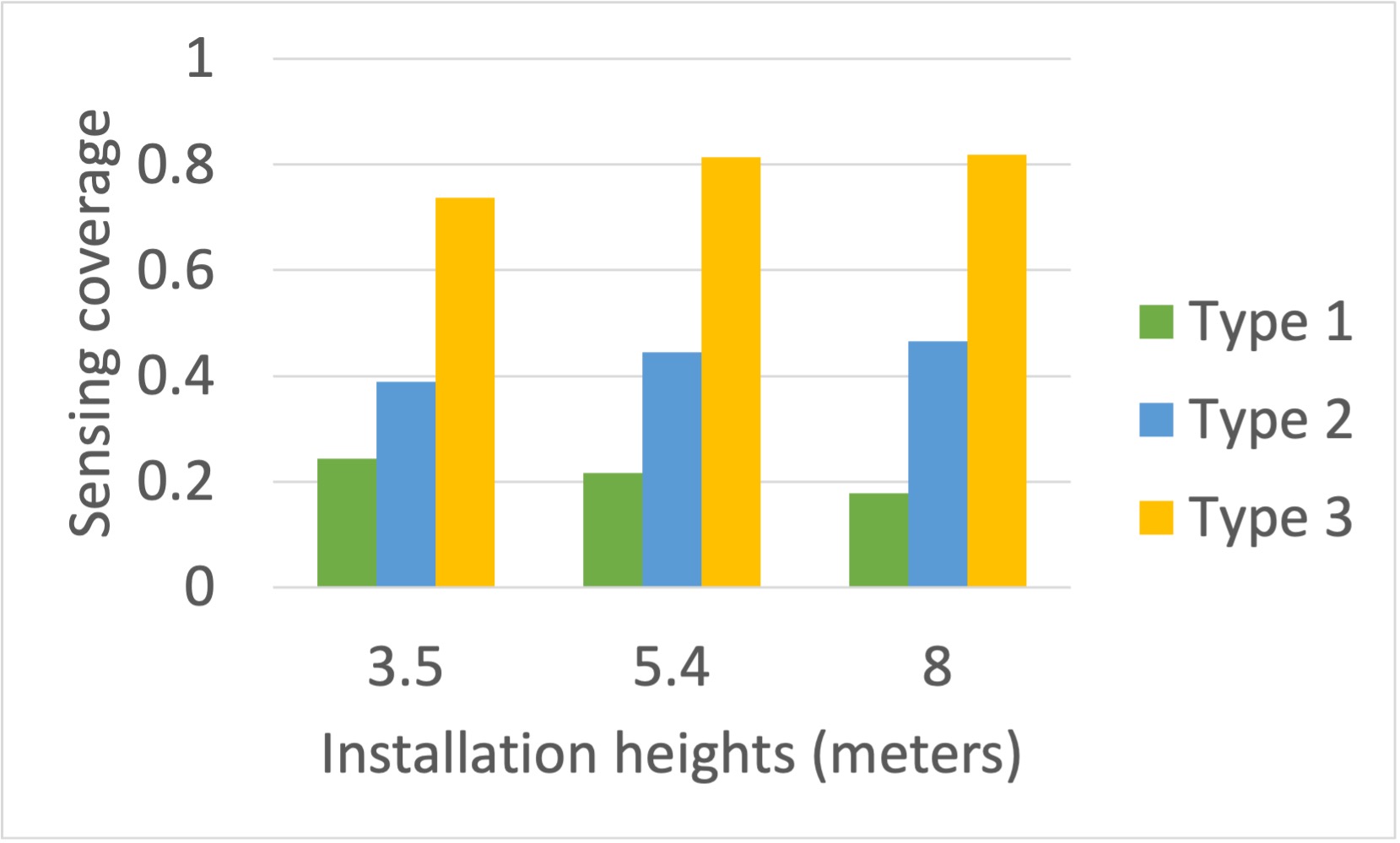}
  \caption{The number of LiDARs=1}
  \label{fig:height_num1}
\end{subfigure}
\begin{subfigure}{0.32\textwidth}
  \centering
  \includegraphics[width=\textwidth]{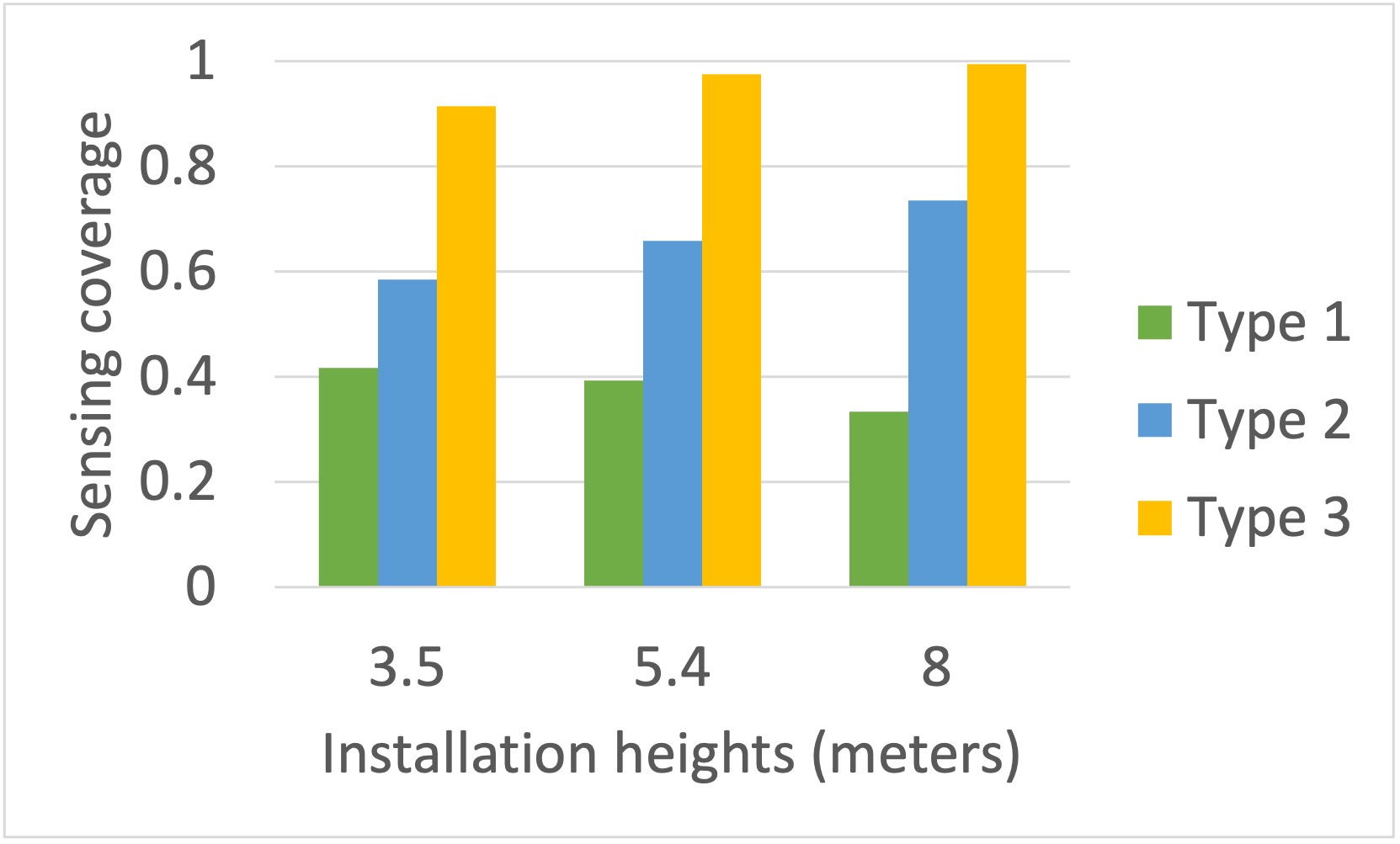}
  \caption{The number of LiDARs=2}
  \label{fig:height_num2}
\end{subfigure}
\begin{subfigure}{0.32\textwidth}
  \centering
  \includegraphics[width=\textwidth]{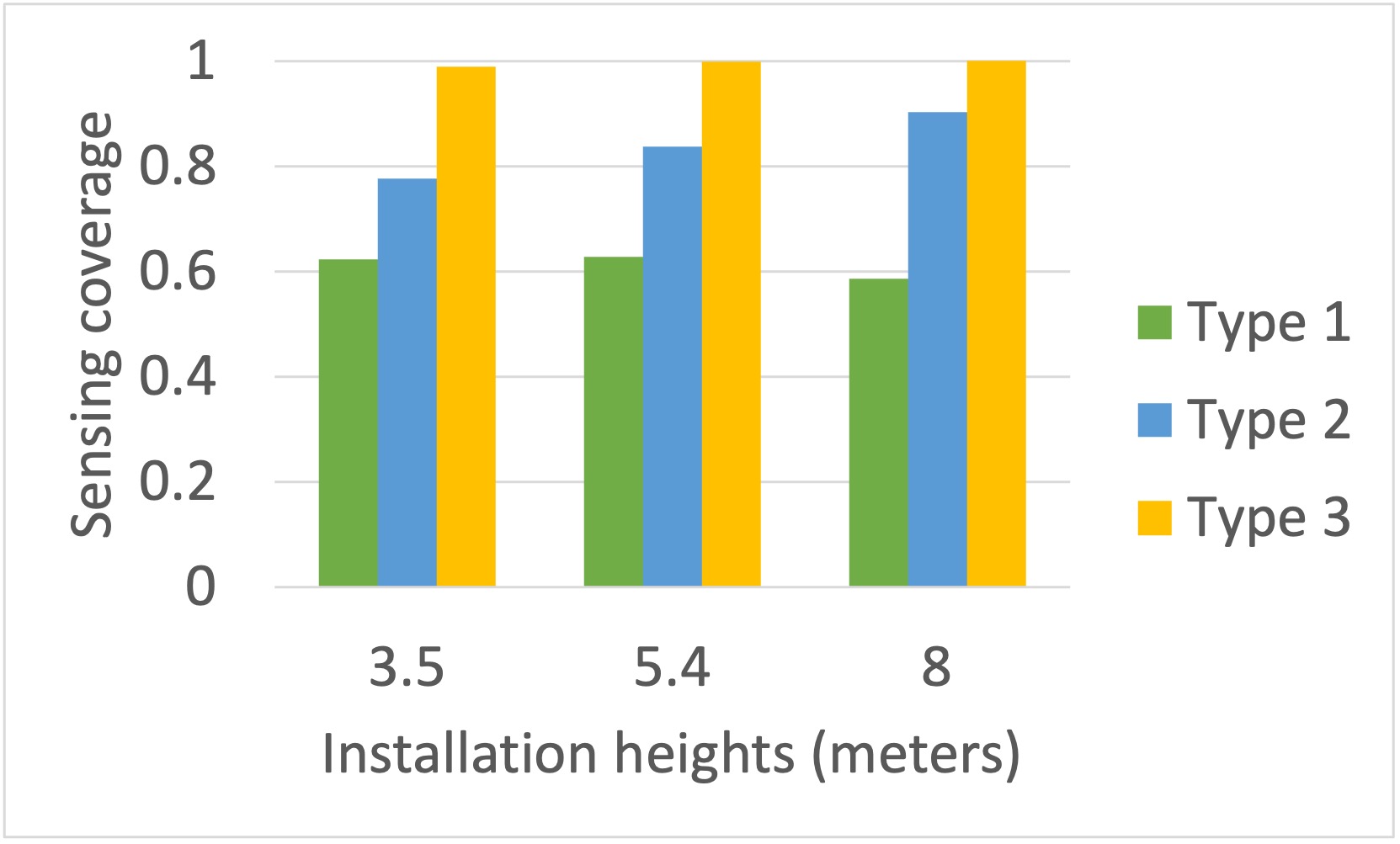}
  \caption{The number of LiDARs=4}
  \label{fig:height_num4}
\end{subfigure}
\caption{The sensing coverage achieved by different LiDAR configurations and heights}
\label{fig:height}
\end{figure*}

To illustrate the usage of the proposed algorithms and simulation-based framework, we use an existing traffic scenario taken from the digital map, titled Town05, provided by the CARLA simulator to construct the synthetic traffic scenario. By deploying infrastructure LiDARs on the roadside, we are interested in detecting vehicles entering and leaving the intersection shown in Fig.~\ref{fig:worldmap}. 

Fig.~\ref{fig:birdview} provides the bird's view of the intersection to monitor. The blue region represents the ROI, while the purple region indicates potential locations where roadside sensors can be deployed to monitor traffic within the ROI. Without loss of generality, we also assume that sensors can be installed at a height within the range between 3.5 and 8 meters, as shown in Fig.~\ref{fig:horizontalview}. This is close to the installation height of traffic lights or roadside poles in urban areas. 

Another assumption we made is related to sensor configurations. Specifically, we operate under the assumption that the sensor deployment engine might choose from the three types of LiDARs currently available in the market. Their specifications are given in Table~\ref{tab:sensor_config}. 

In dealing with object detection tasks downstream of the evaluation pipeline, we use PointPillars \cite{Lang_2019}, the state-of-the-art deep neural network for 3D object detection, to achieve a balance between detection accuracy and execution time. Note that our goal is not to compare the performance gap between different object detectors but the gap between different sensor deployment configurations by using the same object detector (i.e., PointPillar). The performance of the detector is determined by the correctness of the prediction (true positive) if the Intersection-over-Union (IoU), the overlap between the two boxes from a bird's-eye view, is greater than a given threshold. For the vehicle locations predicted by PointPillar, we adopt Average Precision (AP) at an IoU threshold of 0.7 as the metric for performance measurement.

We also suggest that readers take a cautious view of the interpretation of the results we will present next. Though promising, the results only show the possibility of achieving higher marginal gains by adding more low-resolution LiDARs than by adding more high-resolution LiDARs without attesting to every combination of traffic scenarios in the real world. Being aware that road geometry, layout, and traffic patterns can influence both sensor visibility and object detection, readers may first develop their own digital maps that serve the replica of their real-world projects and use our proposed approach for early concept evaluation.

\subsection{Deployment strategies and sensing coverage}
The recommended sensor deployment locations and heights from the optimization engine are indicated by the yellow dots that are shown in Fig.~\ref{fig:recommended_loc}. As expected, it is possible to achieve significantly high sensing coverage by either adding more sensors or upgrading existing LiDARs with high-end ones. However, the difference becomes negligible for Type-3 LiDARs as we add more. This phenomenon is reflected in the redundant recommended locations for Type-3 LiDARs, as shown in Fig.\ref{fig:recommended_loc}. Three Type-3 LiDARs are very close to each other, indicating that \textbf{Algorithm~\ref{alg:bip}} easily finds a feasible solution with the highest possible sensing coverage (i.e., 100$\%$, the maximum possible value for the objective function) without the need for further adjustment. Hence, achieving a 100$\%$ rate with fewer Type-3 sensors is possible. However, for consistency with other sensor types, we mandate the presence of four sensors. 

In addition to sensor resolution, the difference in the sensing coverage also results from installation heights. For example, Type 1 LiDARs need to be installed at a height of 3.5 meters to reach more target points and thus maintain high sensing coverage, as shown in Fig.~\ref{fig:height}. In contrast, Type-2 and 3 LiDARs favor higher positions of installation, regardless of how many sensors we use. Thanks to the long detection range, when installed at the height of 5.4 or 8 meters, light rays emitted by Type-3 LiDARs have a better chance of reaching more target points, reducing occasions. 

\subsection{Considering different importance between regions}
\begin{figure*}[tb!]
\centering
\begin{subfigure}{0.24\textwidth}
  \centering
  \includegraphics[width=\textwidth]{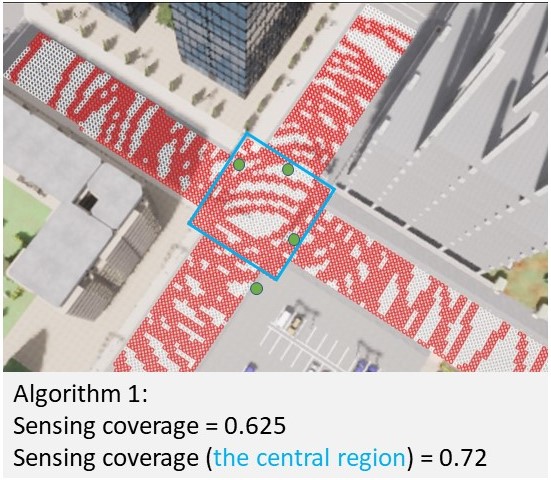}
  \caption{Algorithm 1, Type-1}
  \label{fig:type1_N4}
\end{subfigure}
\begin{subfigure}{0.24\textwidth}
  \centering
  \includegraphics[width=\textwidth]{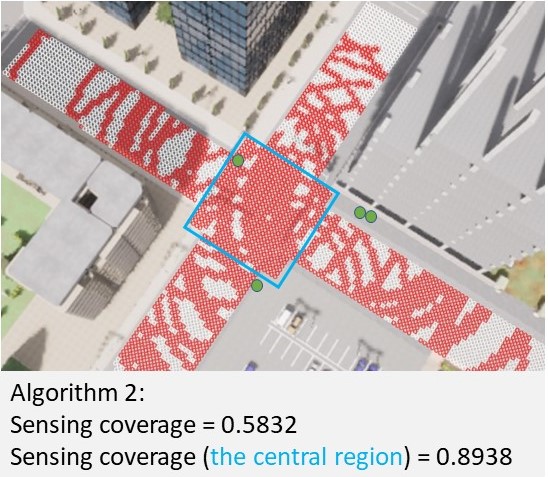}
  \caption{Algorithm 2, Type-1}
  \label{fig:type1_N4_weight}
\end{subfigure}
\begin{subfigure}{0.24\textwidth}
  \centering
  \includegraphics[width=\textwidth]{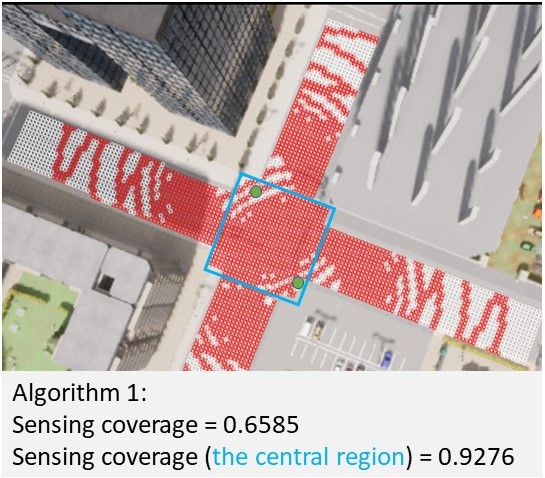}
  \caption{Algorithm 1, Type-2}
  \label{fig:type1_N4}
\end{subfigure}
\begin{subfigure}{0.24\textwidth}
  \centering
  \includegraphics[width=\textwidth]{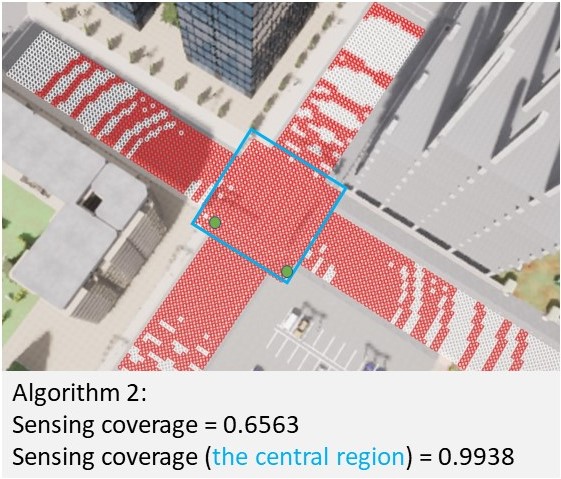}
  \caption{Algorithm 2, Type-2}
  \label{fig:type1_N4_weight}
\end{subfigure}

\caption{The difference in sensing coverage between the vanilla and weighted algorithms}
\label{fig:weight_central}
\end{figure*}

Although Algorithm~\ref{alg:bip} can recommend initial deployment locations and configurations for infrastructure sensors, it does not consider the different levels of importance among ROIs. For example, the central region (marked by blue bounding boxes) in Fig.~\ref{fig:weight_central} might deserve more attention from roadside sensors than other regions near the intersection as conflicts and crashes are more likely to occur among traffic participants. Therefore, with a limited sensing budget, engineers might prefer Algorithm~\ref{alg:bip_weight} that assigns higher priority to the central regions during sensor deployment. In other words, to avoid object detection errors within the central regions, such as the false positives discussed in Section V-C, the sensor deployment strategies should maximize the sensing coverage and point cloud density of the critical region.

Fig.~\ref{fig:weight_central} shows the sensing coverage of the whole intersections and the central region (blue bounding boxes) achieved by Algorithm~\ref{alg:bip} and~\ref{alg:bip_weight}. The green circles represent the deployment locations of sensors, while the red dots mean that the area is visible. For both Type-1 and 2 sensors, Algorithm~\ref{alg:bip_weight} is able to significantly increase the sensing coverage of the central region without compromising the whole coverage. With Type-3 LiDARs, both algorithms achieve almost the same sensing coverage for the central region (the reason we do not include their visibility grids here). This serves to reinforce the necessity of the cost-performance tradeoff in infrastructure sensing projects with budget constraints.

\subsection{The marginal gain by adding more LiDARs}
\begin{figure}[tb!]
\centering
\includegraphics[width=1\columnwidth]{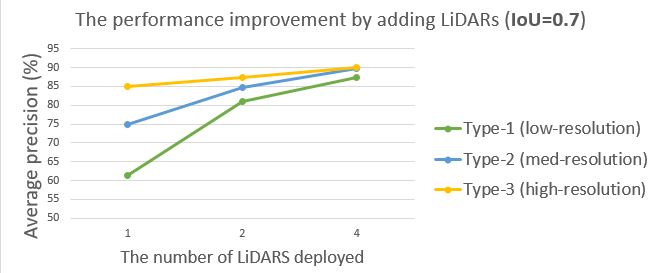}
\caption{Detection accuracy improvement by increasing budget}
\label{fig:det_accuracy}
\end{figure}

Although merging data from multiple infrastructure LiDARS might improve detection accuracy, we also need to consider the cost incurred by adding more sensors. The results from our tests on synthetic datasets suggest that sensor deployment strategies that adopt multiple high-end LiDARs is not necessarily favorable to projects with budget constraints. In particular, adding more low-resolution LiDARS can result in substantially higher performance gains than adding high-resolution ones, as shown in Fig.~\ref{fig:det_accuracy}.

For example, for Type 1 sensors, an approximately 20$\%$ improvement in AP ( at IoU of 0.7) is obtained by increasing the number of LiDAR from one to two, while only a seven percent difference is observed between two and four LiDAR settings, as given in Fig.~\ref{fig:det_accuracy}. For Type 2 sensors, we observed a 10$\%$ improvement by adding another LiDAR to the single-sensor configuration, while only 5$\%$ improvement is achieved by adding two more LiDARs to the dual-sensor solution.

\begin{figure*}[tb!]
\centering
\begin{subfigure}{0.30\textwidth}
  \centering
  \includegraphics[width=\textwidth]{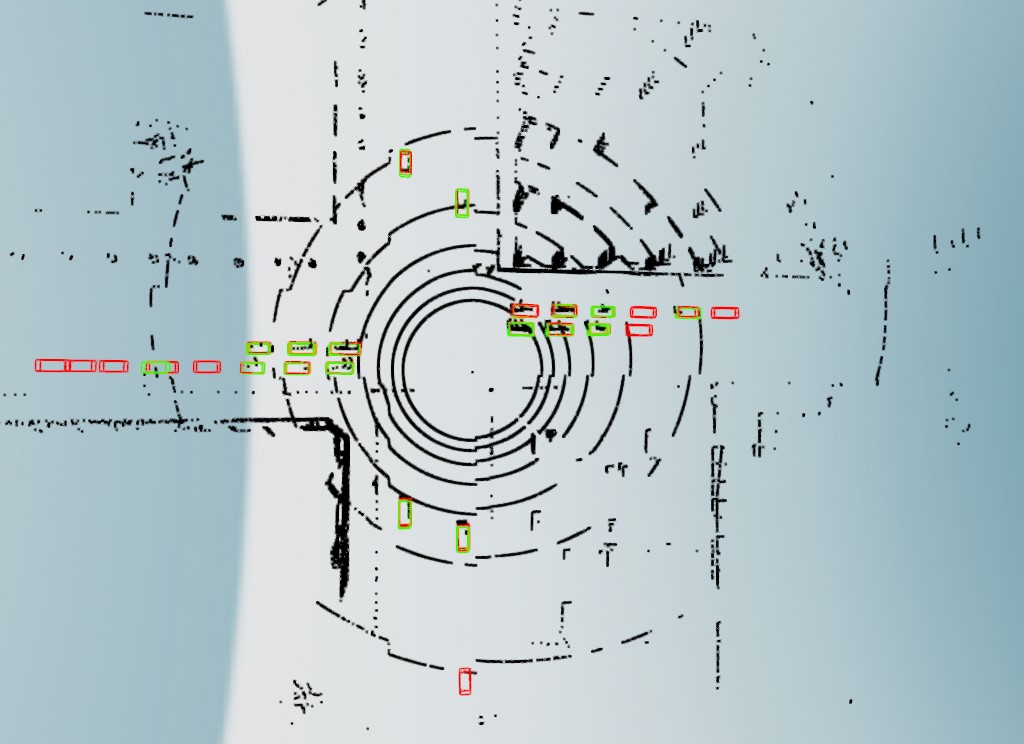}
  \caption{The number of LiDARs=1}
  \label{fig:type1_N1_error}
\end{subfigure}
\begin{subfigure}{0.30\textwidth}
  \centering
  \includegraphics[width=\textwidth]{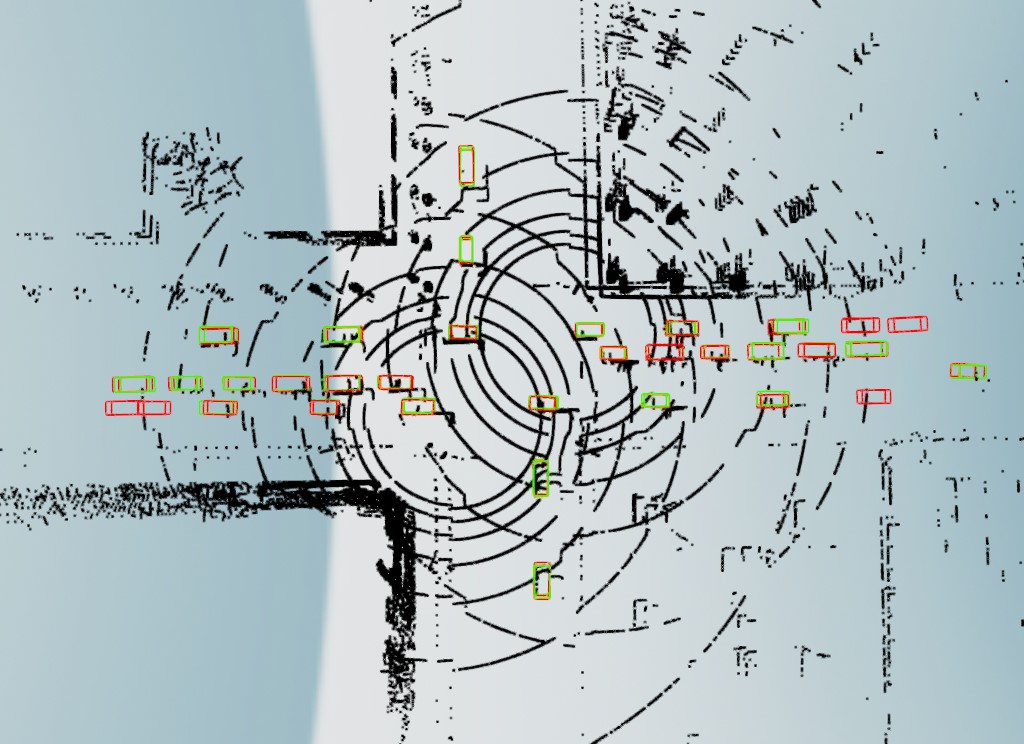}
  \caption{The number of LiDARs=2}
  \label{fig:type1_N2_error}
\end{subfigure}
\begin{subfigure}{0.30\textwidth}
  \centering
  \includegraphics[width=\textwidth]{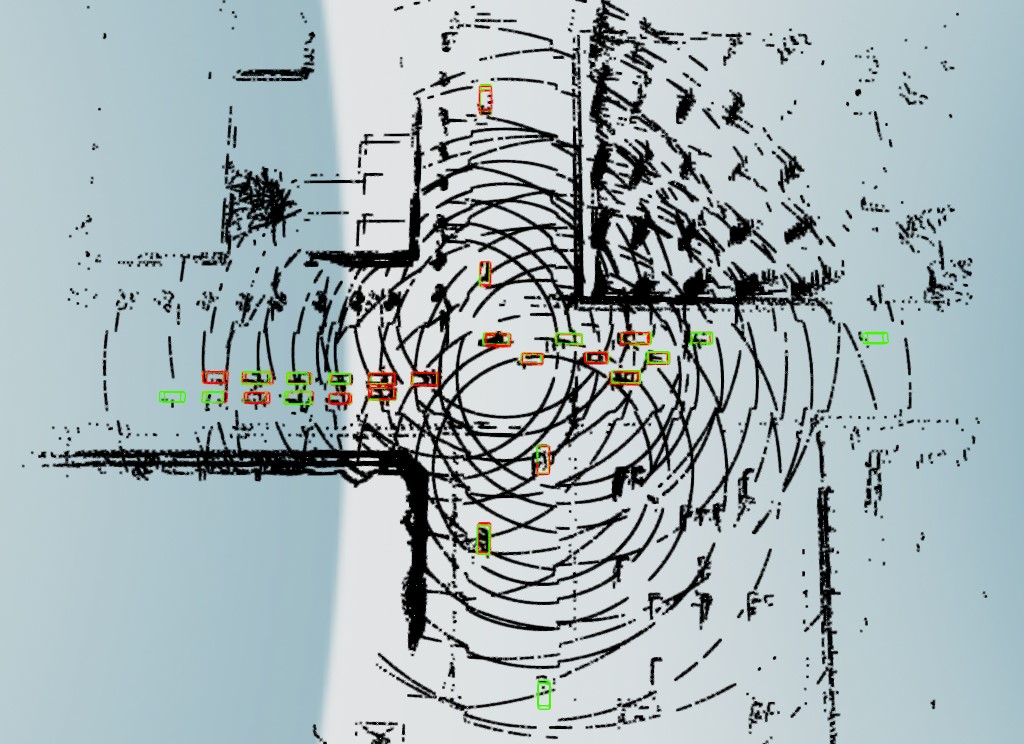}
  \caption{The number of LiDARs=4}
  \label{fig:type1_N4_error}
\end{subfigure}
\caption{The visualization of the detection errors by Type-1 LiDARs}
\label{fig:error}
\end{figure*}

\subsection{Error analysis}
\begin{figure*}[tb!]
\centering
\includegraphics[width=0.7\textwidth]{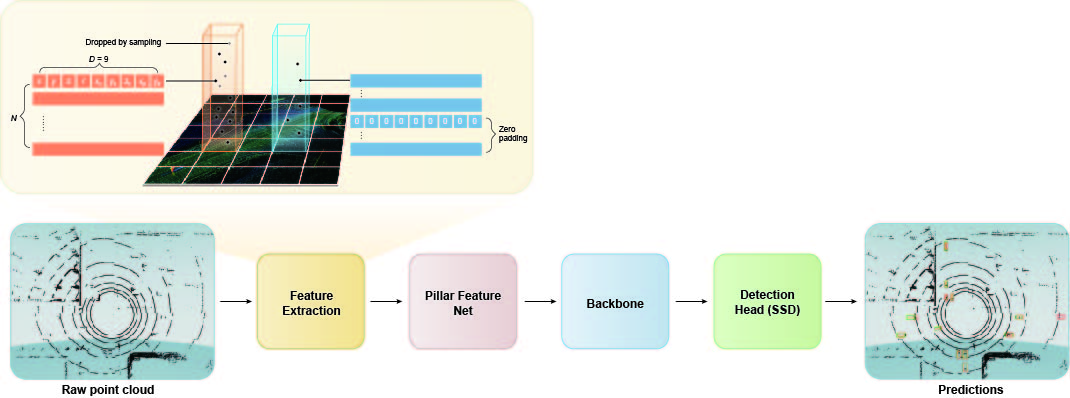}
\caption{The encoding operation within Pointpillar~\cite{Lang_2019}}
\label{fig:pointpillar}
\end{figure*}

Since adding more low-resolution (i.e., Type-1) LiDARs leads to the largest increase in detection performance, it will be beneficial for us to understand the possible reasons behind the improvement. 

For Type-1 LiDARs, the deployment solution with a small sensing budget results in more false positives than the solution with a large sensing budget. For example, the object detection algorithm that uses a single or two LiDARs predicts more vehicles (red boxes) that do not exist in the ground truth (green boxes) than using four LiDARs, as shown in Fig.~\ref{fig:type1_N1_error},~\ref{fig:type1_N2_error}, and~\ref{fig:type1_N4_error}.

When only one sensor is used, we can observe from Fig.~\ref{fig:type1_N1_error} and~\ref{fig:type1_N2_error} that error predictions mainly occur at the boundary regions where the point cloud is sparse. Our interpretation is that the unique operations for point cloud compression within the deep neural network we used (i.e., PointPillar) are prone to mistakes caused by point cloud sparsity. Specifically, for the encoder network that converts the point cloud to a sparse pseudo-image, it will first discretize the cloud into an evenly-spaced grid to create a set of pillars, as shown in Fig.~\ref{fig:pointpillar}. During the encoding process, since all pillars need to convert to 3d tensors with the same size (we call it data augmentation), those pillars with too many points will be randomly sampled, while those with few points will be populated with zero-padding. As a result of the data augmentation, both types of pillars can be detected by the neural network as objects of interest. The problem is that the pillars located near the LiDAR's boundary regions are more likely to contain sparse points that are caused by noises rather than the reflections from real vehicles. The results also imply that the detection range of LiDARs, which provide spatially discretized scans of traffic scenes, is not necessarily a good indication of its actual performance related to traffic monitoring.

\section{CONCLUSIONS}
This paper presents integer programming algorithms and a simulation tool to support the concept evaluation of infrastructure-based collective perception applications. Although preliminary findings in the case study suggest the possibility of achieving higher marginal gains by adding more low-resolution LiDARs, the results call for a cautious and nuanced interpretation, as road geometry, layout, and traffic patterns can influence both sensor visibility and object detection. Future work will explore the use of digital maps built from real-world traffic scenarios in the evaluation process. Additionally, we will explore the influence of deployment uncertainties (e.g., perturbations in sensor location and orientation) on the performance metrics. 






\bibliographystyle{IEEEtran}
\bibliography{root.bib}

\end{document}